\documentclass[11pt,a4paper]{article}
\usepackage[hyperref]{acl2020}
\usepackage{times}
\usepackage{latexsym}
\usepackage{caption}
\usepackage{amsfonts}
\usepackage{amsmath}
\usepackage{graphicx, subfig}
\usepackage{booktabs}
\usepackage{amssymb}
\usepackage{eufrak}
\usepackage{mathalfa}
\usepackage{makecell}
\usepackage{lipsum}
\usepackage{marvosym}
\usepackage{algorithmic,algorithm}
\usepackage{multirow}

\usepackage{multicol} 

\usepackage{microtype}

\aclfinalcopy 
\def\aclpaperid{1361} 

\title{Noise Stability Regularization for Improving BERT Fine-tuning}

\author{Hang Hua$^{1}$\textsuperscript{*\dag},\  Xingjian Li$^{2,3}$\textsuperscript{*},\  Dejing Dou$^{2\textsuperscript{\Letter}}$,\  Chengzhong Xu$^{3}$,\  Jiebo Luo$^1$\\
\\
{ $^1$University of Rochester, Rochester, NY, USA}\\
\texttt{{\{hhua2,\ jluo\}@cs.rochester.edu}}\\
{ $^2$Big Data Lab, Baidu Research, Beijing, China}\\
\texttt{\{lixingjian,\ doudejing\}@baidu.com}\\
{ $^3$Department of Computer Science, University of Macau, Macau, China}\\
\texttt{czxu@um.edu.mo}}

\date{}
\DeclareFontFamily{OT1}{pzc}{}
\DeclareFontShape{OT1}{pzc}{m}{it}{<-> s * [1.10] pzcmi7t}{}
\DeclareMathAlphabet{\mathpzc}{OT1}{pzc}{m}{it}
\begin{document}

\maketitle

{{
{\renewcommand{\thefootnote}{}\footnotetext {\dag Contribution during internship at Baidu Research.}}
{\renewcommand{\thefootnote}{}\footnotetext{* Equal contributions. \Letter ~Correspondence.}}}}

\begin{abstract}
Fine-tuning pre-trained language models such as BERT has become a common practice  dominating leaderboards across various NLP tasks. Despite its recent success and wide adoption, this process is unstable when there are only a small number of training samples available. The brittleness of this process is often reflected by the sensitivity to random seeds. In this paper, we propose to tackle this problem based on the noise stability property of deep nets, which is investigated in recent literature~\cite{Arora2018StrongerGB,Sanyal2020StableRN}. 
Specifically, we introduce a novel and effective regularization method to improve fine-tuning on NLP tasks, referred to as \textbf{L}ayer-wise \textbf{N}oise \textbf{S}tability \textbf{R}egularization (\textbf{LNSR}). We extend the theories about adding noise to the input and prove that our method gives a stabler regularization effect. We provide supportive evidence by experimentally confirming that well-performing models show a low sensitivity to noise and fine-tuning with LNSR exhibits clearly higher generalizability and stability. Furthermore, our method also demonstrates advantages over other state-of-the-art algorithms including $\text{L}^2$-SP~\cite{Li2018ExplicitIB}, Mixout~\cite{Lee2020MixoutER} and SMART~\cite{Jiang2020SMARTRA}. 

\end{abstract}

\section{Introduction}
Large-scale pre-trained language models such as BERT \cite{Devlin2019BERTPO} have been widely used in natural language processing tasks \cite{Guu2020REALMRL,Liu2019FinetuneBF,Wadden2019EntityRA,Zhu2020IncorporatingBI}. A typical process of training a supervised downstream dataset is to fine-tune a pre-trained model for a few epochs.
In this process, most of the model's parameters are reused, while a random initialized task-specific layer is added to adapt the model to the new task. 
\begin{figure*}[t!]
  \centering
  \includegraphics[width=0.85\linewidth]{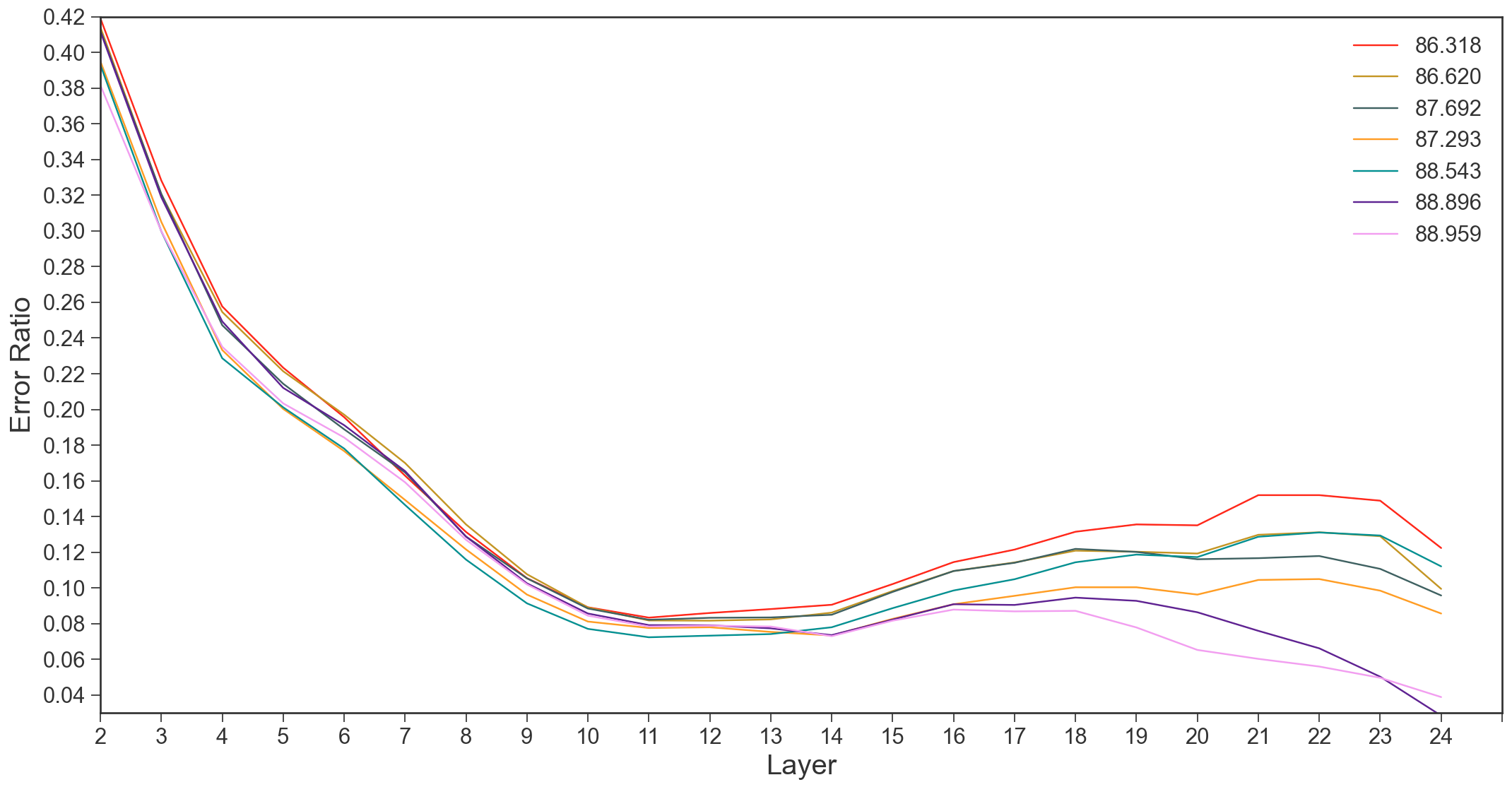}
  \caption{Attenuation of injected noise on the BERT-Large-Uncased model on the MRPC task (X-axis: the layer index. Y-axis: the $L^2$ norm between the original output and noise perturbed output). A curve starts at the layer where a scaled Gaussian noise is injected to its output whose $l_2$ norm is set to $5\%$ of the norm of its original output. As it propagates up, the injected noise has a rapidly decreasing effect on the lower layers but becomes volatile on the higher layers, which indicates the poor generalizability and brittleness of the BERT top layers. Moreover, models with higher accuracies (marked in the upper right) usually have lower error ratios or higher noise stability in top layers.}
  \label{noise}
\end{figure*}

Fine-tuning BERT has significantly boosted the state of the art performance on natural language understanding (NLU) benchmarks such as
GLUE~\cite{Wang2018GLUEAM} and SuperGLUE~\cite{Wang2019SuperGLUEAS}. However, despite the impressive empirical results, this process remains unstable due to the randomness involved by data shuffling and the initialization of the task-specific layer. 
The observed instability in  fine-tuning BERT was first discovered by \citet{Devlin2019BERTPO,Dodge2020FineTuningPL}, and several approaches have been proposed to solve this problem \cite{Lee2020MixoutER,Zhang2020RevisitingFB,Mosbach2020OnTS}.

In this study, we consider the fine-tuning stability  of BERT from the perspective of the sensitivity to input perturbation. This is motivated by \citet{Arora2018StrongerGB} and \citet{Sanyal2020StableRN} who show that noise injected at the lower layers has very little effect on the higher layers for neural networks with good generalizability. However, for a well pre-trained BERT, we find that the higher layers are still very sensitive to the lower layer's perturbation (as shown in Figure~\ref{noise}), 
implying that the high level representations of the pre-trained BERT may not generalize well on downstreaming tasks and consequently lead to instability. This phenomenon coincides with the observation that transferring the top pre-trained layers of BERT slows down learning and hurts performance \cite{Zhang2020RevisitingFB}. In addition, \citet{Yosinski2014HowTA} also point out that in transfer learning models for object recognition, the lower pre-trained layers learn more general features while the higher layers closer to the output specialize more to the pre-training tasks.  We 
argue that this result also applies to BERT. 
Intuitively, if a trained model is insensitive to the perturbation of the lower layers' output, then the model is confident about the output, and vice versa.
Based on the above theoretical and empirical results, we propose a simple and effective regularization method to reduce the noise sensitivity of BERT and 
thus improve the stability and performance of fine-tuned BERT. 

To verify our approach, we conduct extensive experiments on different few-sample (fewer than 10k training samples) NLP tasks, including CoLA \cite{Warstadt2019NeuralNA}, MRPC \cite{Dolan2005AutomaticallyCA}, RTE \cite{Wang2018GLUEAM,Dagan2005ThePR,BarHaim2006TheSP,Giampiccolo2007TheTP}, and STS-B \cite{Cer2017SemEval2017T1}. With the layer-wise noise stability regularization, we obtain strong empirical performance. Compared with other state-of-the-art models, our approach not only improves the  fine-tuning stability (with a smaller standard deviation) but also consistently improve the overall performance (with a larger mean, median and maximum).


In summary, our main contributions are:
\begin{itemize}
\item We propose a lightweight and effective regularization method, referred to as Layer-wise Noise Stability Regularization (LNSR) to improve the local Lipschitz continuity of each BERT layer and thus ensure the smoothness of the whole model. The empirical results show that the fine-tuned BERT models regularized with LNSR obtain significantly more accurate and stable results. LNSR also outperforms other state-of-the-art methods aiming at stabilizing fine-tuning such as $\text{L}^2$-SP~\cite{Li2018ExplicitIB}, Mixout~\cite{Lee2020MixoutER} and SMART~\cite{Jiang2020SMARTRA}.

\item We are the first to study the effect of noise stability in NLP tasks. We extend classic theories of adding noise to explicitly constraining the output consistency when adding noise to the input. We theoretically prove that our proposed layer-wise noise stability regularizer is equivalent to a special case of the Tikhonov regularizer, which serves as a stabler regularizer than simply adding noise to the input~\cite{rifai2011adding}.

\item We investigate the relation of the noise stability property to the generalizability of BERT. We find that in general, models with good generalizability tend to be insensitive to noise perturbation; the lower layers of BERT show a better error resilience property but the higher layers of BERT remain sensitive to the lower layers' perturbation (as is depicted in Figure~\ref{noise}). 
\end{itemize}
\section{Related Work}

\subsection{Pre-training}
Pre-training has been well studied in machine learning and natural language processing \cite{Erhan2009TheDO,Erhan2010WhyDU}. \citet{Mikolov2013DistributedRO} and \citet{Pennington2014GloveGV} proposed to use distributional
representations (i.e., word embeddings) for individual words. \citet{Dai2015SemisupervisedSL} proposed to train
a language model or an auto-encoder with unlabeled data and then leveraged the obtained model
to finetune downstream tasks. Recently, pre-trained language models, like ELMo \cite{Peters2018DeepCW}, GPT/GPT-2 \cite{Radford2018ImprovingLU,Radford2019LanguageMA}, BERT \cite{Devlin2019BERTPO}, cross-lingual language model (briefly, XLM) \cite{Lample2019CrosslingualLM}, XLNet \cite{Yang2019XLNetGA}, RoBERTa \cite{Liu2019RoBERTaAR} and ALBERT\cite{Lan2020ALBERTAL} have attracted more and
more attention in natural language processing communities. The models are
first pre-trained on large amount of unlabeled data to capture rich representations of the input, and
then applied to the downstream tasks by either providing context-aware embeddings of an input sequence \cite{Peters2018DeepCW}, or initializing the parameters of the downstream model \cite{Devlin2019BERTPO} for fine-tuning. Such pre-training approaches deliver decent performance on natural language understanding tasks. 

\subsection{Instability in Fine-tuning }
Fine-tuning instability of BERT has been reported in various previous works. \citet{Devlin2019BERTPO} report instabilities when fine-tuning BERT on small datasets and resort to performing multiple restarts of fine-tuning and selecting the model that performs best on the development set. \citet{Dodge2020FineTuningPL} performs a large-scale empirical investigation of the fine-tuning instability of
BERT. They found dramatic variations in fine-tuning accuracy across multiple restarts and argue how
it might be related to the choice of random seed and the dataset size. \citet{Lee2020MixoutER} propose a new regularization method named Mixout to improve the stability and performance of fine-tuning BERT. \citet{Zhang2020RevisitingFB} evaluate the importance of debiasing step empirically by fine-tuning BERT with both BERTAdam and standard Adam optimizer \cite{Kingma2015AdamAM} and propose a re-initialization method to get a better initialization point for fine-tuning optimization. \citet{Mosbach2020OnTS} analyses the cause of fine-tuning instability and propose a simple but strong baseline (small learning rate combined with bias correction).

\subsection{Regularization}
There has been several regularization approaches to stabilizing the performance of models. \citet{Loshchilov2019DecoupledWD} propose a decoupled weight decay regularizer integrated in Adam \cite{Kingma2015AdamAM} optimizer to prevent neural networks from being too complicate.  \citet{Gunel2020SupervisedCL} use contrastive learning method to augment training set to improve the generalization performance. In addition, spectral norm \cite{Yoshida2017SpectralNR,Roth2019AdversarialTG} serves as a general method can also be used to constrain the Lipschitz continuous of matrix, which can increase the stability of generalized neural networks. 

There are also several noise-based methods have been proposed to improve the generalizability of pre-trained language models, including SMART \cite{Jiang2020SMARTRA}, FreeLB \cite{Zhu2020FreeLBEA} and R3F \cite{Aghajanyan2020BetterFB}. They achieves state of the art performance on GLUE, SNLI \cite{Bowman2015ALA}, SciTail \cite{Khot2018SciTaiLAT}, and ANLI \cite{nie-etal-2020-adversarial} NLU benchmarks. Most of these algorithms employ adversarial training method to improve the robustness of language model fine-tuing. SMART uses an adversarial methodology to encourage models to be smooth within a neighborhoods of all the inputs; FreeLB optimizes a direct adversarial loss through iterative gradient ascent steps; R3F removes the adversarial nature of SMART and optimize the smoothness of the whole model directly. Different from these methods, our proposed method does not adopt the adversarial training strategy, we optimize the smoothness of each layer of BERT directly and thus improve the stability of whole model. 



\section{Using Noise Stability as a Regularizer}
One of the central issues in neural network training is to determine the optimal degree of
complexity for the model. A model which is too limited will not sufficiently capture the structure in the data, while one which is too complex will model the noise on the data (the phenomenon of over-fitting). In either case, the performance on new data, that
is the ability of the network to generalize, will be poor. The problem can be regarded as one of finding the optimal trade-off between the high bias of a model which is too inflexible and the high variance of a model with too much freedom \cite{geman1992neural,bishop1995training,Novak2018SensitivityAG,bishop1991improving}. To control the trade-off of bias against variance of BERT models, we impose an explicit noise regularization method.
\subsection{Introduction of Our Method}
Denoting the training set as $D$, we give the general form of optimization objective for a BERT model $f(\cdot;\theta)$ with $L$ layers, as following:
\begin{equation}
    \theta^*=\mathop{\arg\min}_{\mathbf{\theta}} \underset{(x,y) \sim D}{\mathbb{E}}[ \mathcal{L}(f(x;\theta),y)+ \hat{\mathcal{R}}(\theta)].
\end{equation}

To represent $\hat{\mathcal{R}}(\theta)$, we first define the injection position as the \emph{input} of layer $b$ which is denoted as $x^b$. If the regularization is operated at the \emph{output} of layer $r$, we can further denote the function between layer $b$ and $r$ as $f^{b,r}$, satisfying that $1 <= b <= r <= L$. To implement the noise stability regularization, we inject a Gaussian-like noise vector $\varepsilon$ to $x^b$ and get a neighborhood $x^b + \varepsilon$. Specifically, each element $\varepsilon_i$ is independently randomly sampled from a Gaussian distribution with the mean of zero and the standard deviation of $\sigma$ as $\varepsilon_i \sim \mathcal{N}(0, \sigma^2)$. The probability density function of the noise distribution can be written as $p(\varepsilon_i) = \frac{1}{\sqrt{2\pi} \sigma}e^{-\frac{\varepsilon_i^2}{2\sigma^2}}$. 
Our goal is to minimize the discrepancy between their outputs over $f^{b,r}$ defined as $||f^{b,r}(x^b+\varepsilon)-f^{b,r}(x^b)||^2$. In our framework, we use a fixed position $b$ as the position of noise injection and constrain the output distance on all layers following layer $b$. Denoting the regularization weight corresponding to each $f^{b,r}$ as $\lambda^{b,r}$, given a sample $(x,y) \sim D$, the regularization term is represented by the following formulas:
\begin{equation}
    \hat{\mathcal{R}}(\theta) = \sum_{r=b}^L \lambda^{b,r} ||f^{b,r}(x^b+\varepsilon)-f^{b,r}(x^b)||^2.
\end{equation}
An overall algorithm is represented in Algorithm~\ref{algo:frame}.

\subsection{Theoretical Analysis}
Regularzation is a kind of commonly used techniques to reduce the function complexity and, as a result, to make the learned model generalize well on unseen examples. In this part, we theoretically prove that the proposed LNSR algorithm has the effects of encouraging the local Lipschitz continuity and imposing a Tikhonov regularizer under different assumptions. For simplicity, we omit the notations about the layer number in this part, denoting $f$ as the target function and $x$ as the input of $f$ parameterized by $\theta$. Given a sample $(x, y) \sim D$, we discuss the general form of the noise stability defined as following:
\begin{equation}
   \mathcal{R}(\theta) = \underset{\varepsilon}{E}\{ \|f(x + \varepsilon) - f(x)\|^2 \}.
\label{eq:general}
\end{equation}

\textbf{Lipschitz continuity}. The Lipschitz property reflects the degree of smoothness for a function. Recent theoretical studies on deep learning has revealed the close connection between Lipschitz property and generalization~\cite{bartlett2017spectrally,neyshabur2017exploring}.

Given a sampled $\varepsilon$, minimizing $\|f(x + \varepsilon) - f(x)\|^2$ is equivalent to minimizing:
\begin{equation}
\frac{ \|f(x + \varepsilon) - f(x)\|^2}{ \|x + \varepsilon - x\|^2}.
\end{equation}
Thus the noise stability regularization can be regarded as minimizing the Lipschitz constant in a local region around the input $x$. 

\textbf{Tikhonov regularizer}. The Tikhonov regularizer~\cite{willoughby1979solutions} involves constraints on the derivatives of the objective function with respect to different orders. For the simplest first-order case, it can be regarded as imposing robustness and shaping a flatter loss surface at input, which makes the learned function smoother.

Assuming that the magnitude of $\varepsilon$ is small, we can expand the first term as a Taylor approximation as:
\begin{equation}
    f(x + \varepsilon) = 
    f(x) + \varepsilon \cdot J_f(x) + 
    \frac{1}{2} \varepsilon^T \cdot H_f(x) \cdot \varepsilon + O(\varepsilon^3),
\label{eq:taylor}
\end{equation}
where $J_f(x)$ and $H_f(x)$ refer to the Jacobian and Hessian of $f$ with respect to the input $x$ respectively. 
 
Ignoring the higher order term $O(\mathbf{\varepsilon}^3)$ and denoting $f_k$ as the k-th output of the function $f$, we can rewrite the regularizer by substituting Eq.~\ref{eq:taylor} in Eq.~\ref{eq:general} as:
\begin{equation} 
\begin{aligned}
    \mathcal{R}(\theta)\! & \!= \!\underset{\mathbf{\varepsilon}}{E}\{ \| \mathbf{\varepsilon} \cdot J_f(x) +
    \frac{1}{2} \mathbf{\varepsilon}^T \cdot H_f(x) \cdot \mathbf{\varepsilon} \|^2 \} \\
    &= \!\int \| \mathbf{\varepsilon}\!\cdot\!
    J_f(x) +
    \frac{1}{2} \mathbf{\varepsilon}^T\!\cdot\! H_f(x)\!\cdot\!\mathbf{\varepsilon} \|^2 p(\mathbf{\varepsilon}) d\mathbf{\varepsilon} \\
    &= \!\sum_{k}\!\int\!\|\mathbf{\varepsilon}\!\cdot\!J_{f_k}(x)\!+\!\frac{1}{2}\! \mathbf{\varepsilon}^T\!\cdot\!H_{f_k}(x) \!\cdot\!\mathbf{\varepsilon} \|^2 p(\mathbf{\varepsilon}) d\mathbf{\varepsilon}
    .
\label{eq:main}
\end{aligned}
\end{equation}

We define the input vector $x$ as $(x_1, x_2, ......, x_{d_1})$ and noise vector $\mathbf{\varepsilon}$ as  $\mathbf{\varepsilon} = (\varepsilon_1, \varepsilon_2, ...... , \varepsilon_{d_1})$. Assuming that distributions of the noise and the input are irrelevant, and the derivative of $f$ with respect to different elements of the input vector is independent with each other, we expand the second order term corresponding to the Jacobian as:
\begin{equation}
\begin{aligned}
    \Omega_J(f_k) &= 
    \int \| \mathbf{\varepsilon} \cdot J_{f_k}(x) \|^2 p(\mathbf{\varepsilon}) d\mathbf{\varepsilon} \\
     &= \int \sum_{i} (\varepsilon_i \frac{\partial f_k}{\partial x_i})^2 p(\mathbf{\varepsilon}) d\mathbf{\varepsilon} \\
     &= \sum_{i} (\frac{\partial f_k}{\partial x_i})^2 \int \varepsilon_i^2 p(\mathbf{\varepsilon}) d\mathbf{\varepsilon} \\ 
     &= \sigma^2 \|J_{f_k}(x)\|^2.
\label{eq:jacobian}
\end{aligned}
\end{equation}

According to the characteristics of the Gaussian distribution, we also have
\begin{equation}
    \int \varepsilon_i \varepsilon_j p(\mathbf{\varepsilon}) d\mathbf{\varepsilon} = 0 \text{\ \ \ for any\ } i \ne j. 
\end{equation}
Thus, we can rewrite the second order term corresponding to the Hessian in Eq.~\ref{eq:main} as:
\begin{equation}
\begin{aligned}
    \Omega_H(f_k) &= \int \| \frac{1}{2} \mathbf{\varepsilon}^T \cdot H_{f_k}(x) \cdot \mathbf{\varepsilon} \|^2 p(\mathbf{\varepsilon}) d\mathbf{\varepsilon} \\
     &= \int \frac{1}{4} \sum_i ( \varepsilon_i^2 \frac{\partial^2 f_k}{\partial x_i^2} )^2 p(\mathbf{\varepsilon}) d\mathbf{\varepsilon} \\
     &= \frac{1}{4} \sum_i (\frac{\partial^2 f_k}{\partial x_i^2})^2 \int \varepsilon_i^4 p(\mathbf{\varepsilon}) d\mathbf{\varepsilon} \\
     &= C \|\mathrm{Tr}(H_{f_k}(x)^T H_{f_k}(x))\|^2,
\label{eq:hessian}
\end{aligned}
\end{equation}
Where $C$ is a constant independent of the input $x$. The third term generated from the expansion of Eq.~\ref{eq:main} is zero as we have $\int \mathbf{\varepsilon}^3 p(\mathbf{\varepsilon}) d\mathbf{\varepsilon} = 0$. Thus we get 
\begin{equation}
    \mathcal{R}(\theta) = \sum_{k}\{ \Omega_J(f_k) + \Omega_H(f_k) \}.
\label{eq:final}
\end{equation}

Considering that the input and output of the function $f$ are both scalar variable, the Tikhonov regularization~\cite{willoughby1979solutions} takes the general form as: 
\begin{equation}
    \mathcal{R_T}(\theta) = \sum_{r} \int h_r(x) (\frac{\partial^r f}{\partial x^r})^2 dx.
\end{equation}
Eq.~\ref{eq:final} shows that our proposed regularizer ensuring the noise stability is equivalent to a special case of the Tikhonov regularizer, where we involve the first and second order derivatives of the objective function $f$. 

An alternative for improving the robustness is to directly add noise to the input, without explicitly constraining the output stability. ~\cite{rifai2011adding} has derived that adding noise to the input has the effect of penalizing both the $L_2$-norm of the Jacobian $\|J_f(x)\|^2$ and the trace of the Hessian $\mathrm{Tr}(H_f(x))$, whereas the Hessian term is not constrained to be positive. While the regularizer brought by our proposed \textbf{LNSR} is guaranteed to be positive by involving the sum of squares of the first and second order derivatives. Moreover, our work relaxes the assumption of MSE regression loss required by~\cite{rifai2011adding}. By imposing the explicit constraint of noise stability on middle layer representations, 
we extend the theoretical understanding of noise stability into deep learning algorithms.

\begin{algorithm}[h]
\caption{Layer-wise Noise Stability Regularization (LNSR)} 
\begin{algorithmic}[1]
\REQUIRE Training set $D$, perturbation bound $\delta$, learning rate $\tau$, number of layers $L$, number of training epochs $N$, function $f$ and its corresponding parameters  $\theta$, the position of noise injection $b$, and 
regularization weights for each layer $\{\lambda^b,...,\lambda^L\}$.
\STATE Initialize $\theta$
\FOR{epoch=$1,2,...,N$ } 
    \FOR{minibatch $B \sim D$}
    \STATE $\hat{\mathcal{R}}\gets 0$
    \FOR{each $x \in B$}
        \STATE $\varepsilon \sim \mathcal{N}(0, \sigma^2)$
        \STATE $\tilde{x}\gets x+\varepsilon$
        \STATE forward pass given $x$ and $\tilde{x}$ as inputs
            \FOR{$r=b,b+1,...,L$}
            \STATE $\hat{\mathcal{R}}\gets \hat{\mathcal{R}}+\lambda^{b,r}||f^{b,r}(x)-f^{b,r}(\tilde{x})||^2$
            \ENDFOR
        \ENDFOR
    \STATE $g \gets \frac{1}{|B|}\sum_{(x,y)}
    \nabla[\mathcal{L}(f(x;\theta),y)+\hat{\mathcal{R}}] $
    \STATE$ \theta\gets \theta-\tau g $
    \ENDFOR
\ENDFOR
\ENSURE $\theta$
\end{algorithmic}
\label{algo:frame}
\end{algorithm}

\begin{center}
\begin{table*}
  \resizebox{\textwidth}{1.95cm}{
  \begin{tabular}{lllllllllllll}
    \toprule
    \multirow{2}{*}{} & \multicolumn{3}{c}{RTE} & \multicolumn{3}{c}{MRPC} & \multicolumn{3}{c}{CoLA}& \multicolumn{3}{c}{STS-B} \\
\cmidrule(r){2-4} \cmidrule(r){5-7} \cmidrule(r){8-10} \cmidrule(r){11-13}
&  mean      &  std   &   max
&  mean      &  std   &   max
&  mean      &  std   &   max 
&  mean      &  std   &   max \\
\midrule
    FT \cite{Devlin2019BERTPO}& $70.13$& $1.84$& $72.56 $& $87.57$&$0.92$&$89.16 $&$60.54 $&$1.49 $& $62.59$ &$89.38 $&$0.53 $& $90.23$ \\
    $\text{L}^2\text{-SP}$ \cite{Li2018ExplicitIB}& $70.58$& $\bf{1.29}$& $73.28$& $ 87.74$&$0.86 $&$88.95 $& $60.19$ &$1.42 $&$63.89 $& $89.25$ & $0.62$&$90.14 $\\
    Mixout\cite{Lee2020MixoutER}& $71.35$& $1.66$& $74.36$& $87.63$&$0.62$&$88.91 $&$63.12 $&$1.68 $& $65.12$ &$89.58 $&$0.35 $& $90.11$ \\
    SMART\cite{Jiang2020SMARTRA}& $72.23$& $2.41$& $75.45$& $87.86$&$0.63$&$89.09 $&$63.16 $&$1.17 $& $65.21$ &$90.11 $&$0.33 $& $90.83$ \\
    \midrule
    \textbf{LNSR(ours)}& $\bf{73.31}$& $1.55$& $\bf{76.17}$& $\bf{88.50}$&$\bf{0.56}$&$\bf{90.02} $&$\bf{63.35} $&$\bf{1.05} $& $\bf{65.99}$ &$\bf{90.23} $&$\bf{0.31} $& $\bf{90.97}$\\
    \bottomrule
  \end{tabular}
  }
    \caption{The mean, standard deviation, and maximum performance on the development set of RTE, MRPC, CoLA, and STS-B tsak across 25 random seeds when fine-tuning the BERT-Large model with various regularization methods. FT refers to the standard BERT fine-tuning. Standard deviation: lower is better.}
    \label{tab:overall}
\end{table*}
\end{center}

\begin{figure*}[htbp]
\centering 
\begin{minipage}[t]{0.49\textwidth}
\centering
\includegraphics[height=2.89cm]{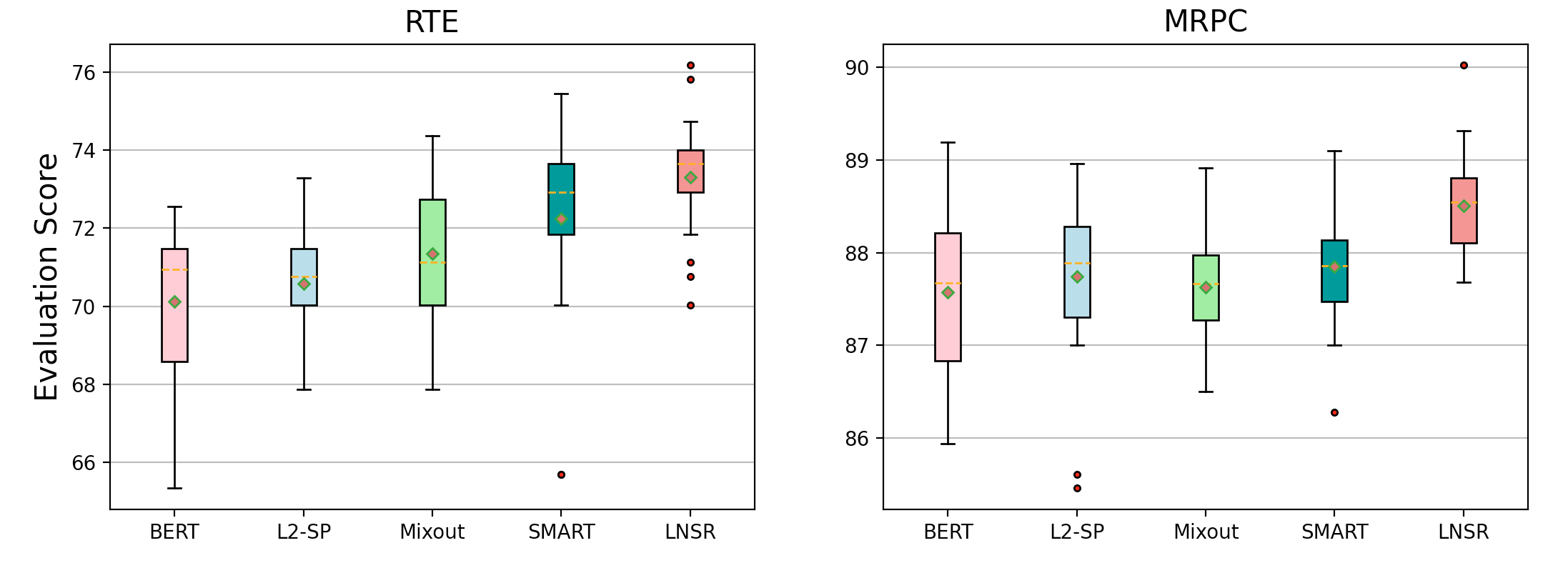}
\end{minipage}
\begin{minipage}[t]{0.49\textwidth}
\centering
\includegraphics[height=2.89cm]{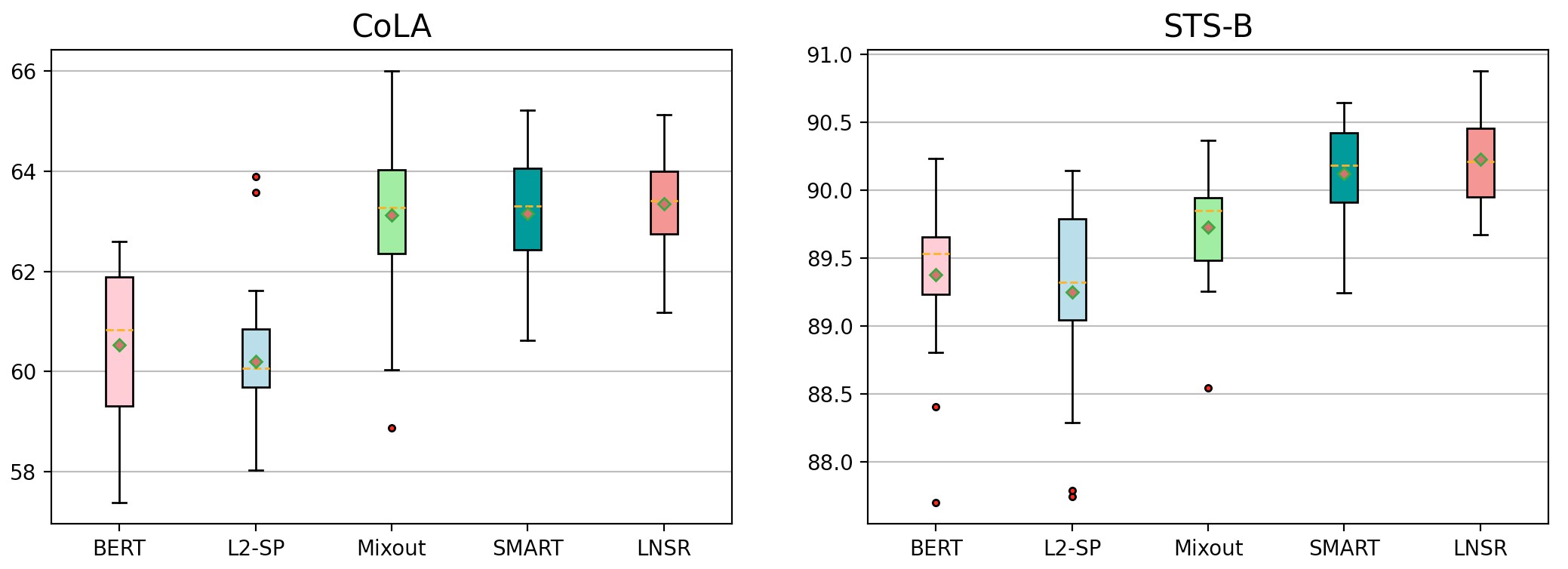}
\end{minipage}
\caption{Performance distribution box plot of each model on the four tasks from 25 random seeds. }
\label{fig:overall}
\end{figure*}
\vspace{-0.4cm}
\section{Experiments}
In this section, we experimentally demostrate the effectiveness of LNSR method on text classification tasks over other regularization methods, and confirm that the insensitivity to noise promotes the generalizability and stability of BERT.

\subsection{Data}
We conduct experiments on four few-sample (less than 10k training samples) text classification tasks of GLUE \footnote{https://gluebenchmark.com/}, the datasets are described below and summarized in Appendix~\ref{sec:experimental} Table~\ref{tab:datasets}. 

\textbf{Corpus of Linguistic Acceptability} (CoLA \cite{Warstadt2019NeuralNA}) consists of English acceptability judgments drawn from books and journal articles on linguistic theory. Each example is a
sequence of words annotated with whether it is a grammatical English sentence. This is a binary classification task and Matthews correlation coefficient (MCC) \cite{matthews1975comparison} is used to evaluate the performance.

\textbf{Microsoft Research Paraphrase Corpus} (MRPC \cite{Dolan2005AutomaticallyCA}) is a corpus of sentence pairs with human annotations for whether
the sentences in the pair are semantically equivalent. The evaluation metrics is the average of F1 and Accuracy.

\textbf{Recognizing Textual Entailment} (RTE \cite{Wang2018GLUEAM}) \cite{Dagan2005ThePR} \cite{BarHaim2006TheSP} \cite{Giampiccolo2007TheTP} is a corpus of textual entailment, and each example is a sentence pair annotated whether the first entails the second. The evaluation metrics is Accuracy.

\textbf{Semantic Textual Similarity Benchmark} (STS-B \cite{Cer2017SemEval2017T1})is a regression task. Each example is a sentence pair and is human-annotated with a similarity score from 1 to 5; the task is to predict these scores. The evaluation metrics is the average of Pearson and Spearman correlation coefficients.

\subsection{Baseline Models}
We use BERT \cite{Devlin2019BERTPO}, a large-scale bidirectional pre-trained language model as the base model in all experiments. We adopt pytorch edition implemented by \citet{Wolf2019HuggingFacesTS}.

\textbf{Fine-tuning}. We use the standard BERT fine-tuning method described in~\citet{Devlin2019BERTPO}.

\textbf{$\text{L}^2$-SP} \cite{Li2018ExplicitIB} is a regularization scheme that explicitly promotes the similarity of the final solution with the initial model. It is usually used for preventing pre-trained models from catastrophic forgetting.  We adopt the form of $\Omega(w)=\frac{\alpha}{2}||w_s-w_s^0||+\frac{\beta}{2}||w_{\bar{s}}||$.

\textbf{Mixout} \cite{Lee2020MixoutER} is a stochastic regularization technique motivated by Dropout \cite{Srivastava2014DropoutAS} and DropConnect \cite{Wan2013RegularizationON}. At each training iteration, each model parameter is replaced with its pre-trained value with probability $p$. The goal is to improve the generalizability of pre-trained language models.

\textbf{SMART} \cite{Jiang2020SMARTRA} imposes an smoothness regularizer inducing an adversarial manner to control the model complexity at the fine-tuning stage. It also employs a class of Bregman proximal point optimization methods to prevent the model from aggressively updating during fine-tuning.

\subsection{Experimental Setup}
Our model is implemented using Pytorch based on Transformers framework \footnote{https://huggingface.co/transformers/index.html}. Specifically, we use the learning setup and hyperparameters recommended by \cite{Devlin2019BERTPO}. We use Huggingface edition Adam \cite{Kingma2015AdamAM} optimizer (without bias correction) with learning rate of $2 \times 10^{-5}$,$\beta_1=0.9, \beta_2=0.999$, and warmup over the first 10\% steps of the total steps. We fine-tune the entire model (340 million parameters), of which the vast majority start as pre-trained weights (BERT-Large-Uncased) and the classification layer (2048 parameters). Weights of the classification layer are initialized with $\mathcal{N}(0,0.02^2)$. We train with a batch size of 32 for 3 epochs. More details of our experimental setup are described in Appendix~\ref{sec:experimental}.

\begin{center}
\begin{table*}
  \label{tab:commands}
  \resizebox{\textwidth}{1.81cm}{
  \begin{tabular}{lllllllllllll}
    \toprule
    \multirow{2}{*}{} & \multicolumn{3}{c}{RTE} & \multicolumn{3}{c}{MRPC} & \multicolumn{3}{c}{CoLA}& \multicolumn{3}{c}{STS-B} \\
\cmidrule(r){2-4} \cmidrule(r){5-7} \cmidrule(r){8-10} \cmidrule(r){11-13}
&  mean      &  std   &   max
&  mean      &  std   &   max
&  mean      &  std   &   max 
&  mean      &  std   &   max \\
\midrule
    FT & $70.13$& $1.84$& $72.56 $& $87.57$&$0.92$&$89.16 $&$61.56 $&$1.34 $& $64.10$ &$89.38 $&$0.53 $& $90.23$ \\
    FT (4 Epochs) & $70.69$& $1.97$& $73.65 $& $88.15$&$0.65$&$89.21 $&$60.69 $&$1.24 $& $62.09$ &$89.29 $&$0.56 $& $90.12$ \\
    FT+Noise & $70.62$& $1.56$& $72.93 $& $87.95$&$0.83$&$89.33 $&$60.18 $&$1.58 $& $62.59$ &$89.34 $&$0.51 $& $90.11$ \\
    \textbf{LNSR}(ours)& $\bf{73.31}$& $\bf{1.55}$& $\bf{76.17}$& $\bf{88.50}$&$\bf{0.56}$&$\bf{90.02} $&$\bf{63.35} $&$\bf{1.05} $& $\bf{65.99}$ &$\bf{90.23} $&$\bf{0.31} $& $\bf{90.97}$\\
    \bottomrule
  \end{tabular}
  }
    \caption{Ablation study of LNSR on each task, we report the mean evaluation scores and standard deviation and max value across 25 random seeds. FT refers to standard BERT fine-tuning.}
    \label{tab:ablation}
\end{table*}
\end{center}

\begin{center}
\begin{table*}
  \label{tab:commands}
  \resizebox{\textwidth}{1.35cm}{
  \begin{tabular}{lllll}
    \toprule
    \multirow{2}{*}{} &
    RTE & MRPC & CoLA & STS-B \\
    \cmidrule(r){2-5}
&  train / eval / gap
&  train / eval / gap
&  train / eval / gap
&  train / eval / gap
\\
\midrule
    FT \cite{Devlin2019BERTPO}& $\bf{95.89}$/$70.13$/$25.76 $& $96.57$/$87.57$/$9.00$&
    $\bf{97.71} $/$61.56 $/$36.25$ &
    $98.31 $/$89.38 $/ $8.93$ \\
    \textbf{LNSR(ours)}& 
    $90.72$/$\bf{73.31}$/$\bf{17.41}$& $\bf{96.68}$/$\bf{88.50}$/$\bf{8.18} $&
    $93.44 $/$\bf{63.35} $/$\bf{30.09}$ &
    $\bf{98.45} $/$\bf{90.23} $/ $\bf{8.22}$\\
    \bottomrule
  \end{tabular}
  }
    \caption{Comparison of the generalizability performance of different models. We report the mean training Acc and evaluation Acc and the generalizability gap (training Acc - evaluation Acc) of each model across 20 random seeds.}
    \label{tab:geng}
\end{table*}
\end{center}

\vspace{-1.4cm}
\subsection{Overall Performance}
Table~\ref{tab:overall} shows the results of all the models on selected GLUE datasets. We train each dataset over 25 random seeds. To implement our LNSR, we uniformly inject noise at the first layer on BERT-large for the comparison with baseline models. As we can see from the table, our model outperforms all the baseline models in mean and max values, which indicates the stronger generalizability of our model against other baseline models.  The p-values between the accuracy distributions of standard BERT fine-tuning and our model are calculated to verify whether the improvements are significant. We obtain very small p-values in all tasks: RTE: $9.7\times10^{-7}$, MRPC: $2.3\times10^{-4}$, CoLA: $4.7\times10^{-8}$, STS-2: $3.3\times10^{-8}$.

Standard deviation is an indicator of the stability of models' performance and 
higher std means more sensitive to random seeds. Our model shows a lower standard deviation on each task, which means our model is less sensitive to random seeds than other models. Figure~\ref{fig:overall} presents a clearer illustration. To sum up, our proposed method can effectively improve the performance and stability of fine-tuning BERT.

\section{Analysis}
\subsection{Ablation Study}
To verify the effectiveness of our proposed LNSR model, we  conduct several ablation experiments including fine-tuning with more training epochs and noise perturbation without regularization (we inject noise directly to the output of a specific layer, and then use the perturbed representation to conduct propagation and then calculate loss, this process is similar to a vector-space represent augmentation). The results 
are shown in Table ~\ref{tab:ablation}. We observe that benefit obtained by longer training is limited. Similarly, fine-tuning with noise perturbation only achieves slightly better results on two of these tasks, showing that simply adding noise without an explicit restriction on outputs may not be sufficient to obtain good generalizability.  
While BERT models with LNSR perform better on each task. This verifies our claim that LNSR can promote the stability of BERT fine-tuning and meanwhile improve the generalizability of the BERT model.

\begin{figure}[t!]
  \centering
  \includegraphics[width=1.0\linewidth]{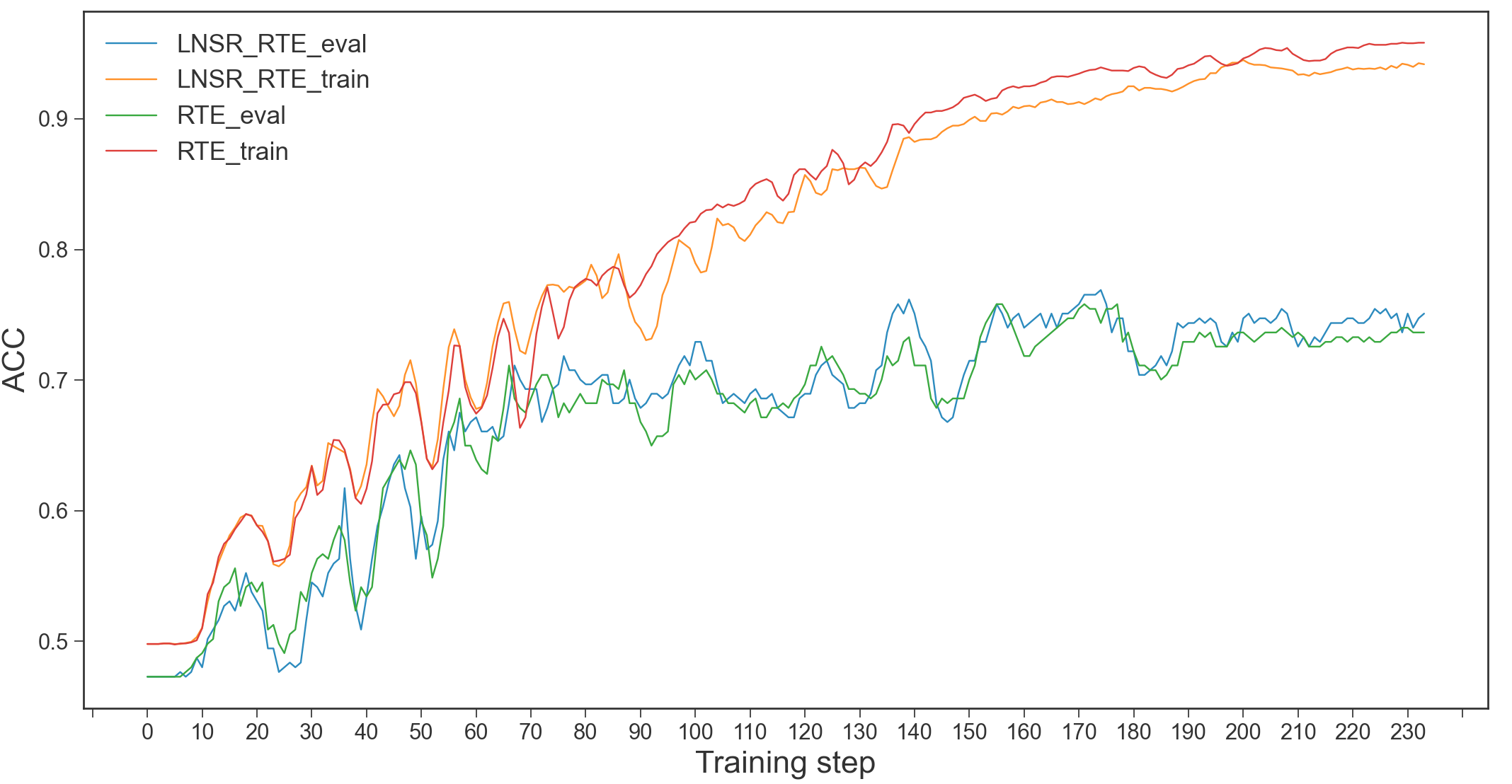}
  \includegraphics[width=1.0\linewidth]{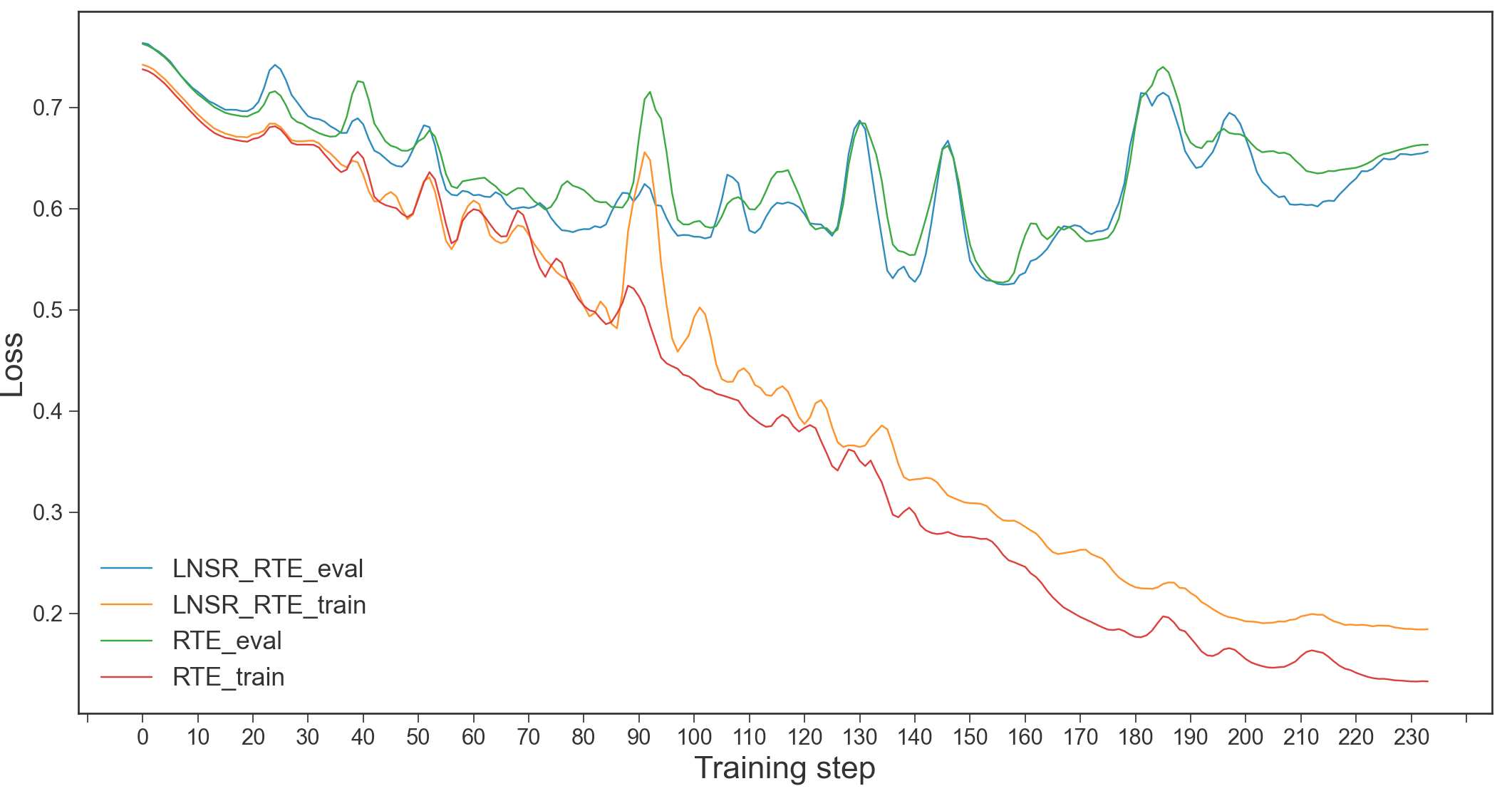}
  \caption{The top-1 accuracy (top) and loss (bottom) curve for fine-tuning BERT with and w/o LNSR.}
  \label{fig:loss}
  \vspace{-0.4cm}
\end{figure}

\subsection{Effects on the Generalizability of Models}
We verify the effects of our proposed method on the generalizability of BERT models in two ways -- generalization gap and models' performance on fewer training samples. Due to the limited data and the extremely high complexity of BERT model, bad fine-tuning start point makes the adapted model overfit the training data and does not generalize well to unseen data. Generalizability of models can be intuitively reflected by generalization gap and models' performance on fewer training samples.

Table~\ref{tab:geng} shows the mean training Acc, mean evaluation Acc and generalization gap of different models on each task. As we can see from the table, fine-tuning with LNSR can effectively narrow the generalization gap, and achieve higher evaluation score. The effect of narrowing generalization gap is also reflected in Figure~\ref{fig:loss} where we can see the higher evaluation accuracy and lower evaluation loss.

\begin{figure}[h!]
  \centering
  \includegraphics[width=0.85\linewidth]{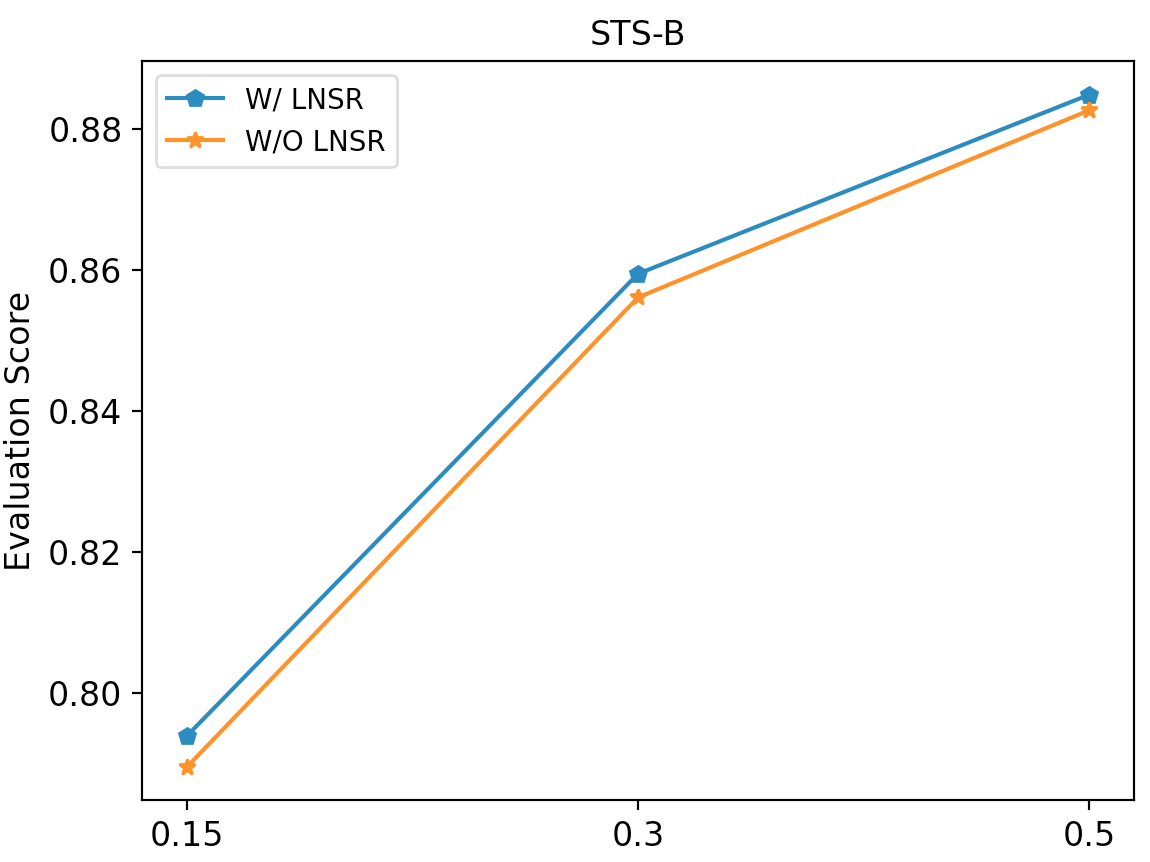}
  \includegraphics[width=0.85\linewidth]{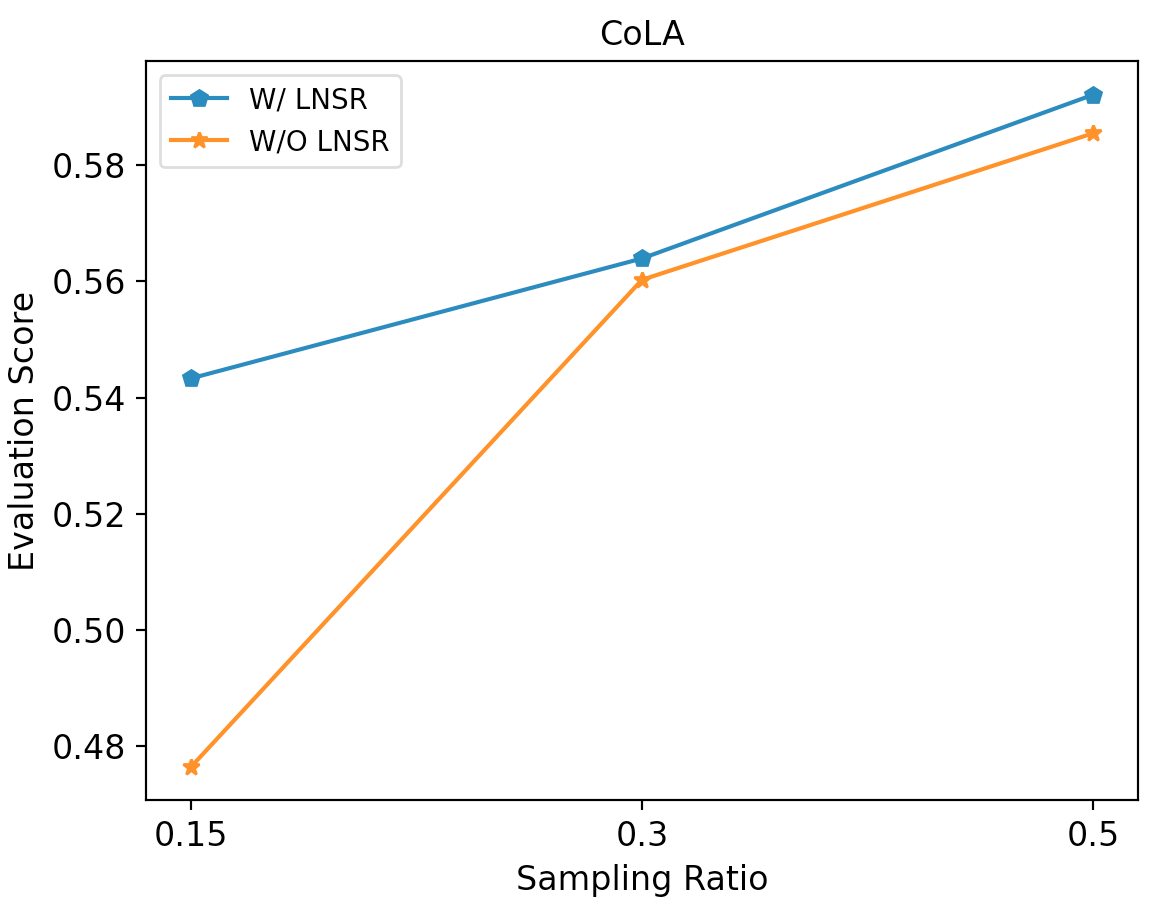}
  \caption{Mean evaluation score comparison of fine-tuning BERT with and w/o LNSR on fewer training samples of the CoLA and STS-B tasks. The mean values are calculated over 20 random seeds.}
  \label{colads}
\end{figure}

We sample subsets from the two relatively larger datasets CoLA (8.5k training samples) and STS-B (7k training samples) with the sampling ratio of 0.15, 0.3 and 0.5.  As is shown in Figure~\ref{colads}, fine-tuning with LNSR shows clear advantage on fewer training samples,  suggesting LNSR can effectively promote the model's generalizability.

\subsection{Sensitivity to the Position of Noise Injection}
We briefly discuss about the sensitivity to the position of noise injection as it is a pre-determined hyperparameter of our method. As is shown in Figure~\ref{fig:sens} in Appendix~\ref{sec:experimental}, we observe that the performance of LNSR does not fluctuate much as the position of noise injection changes. All injection positions bring significant improvements over vanilla fine-tuning. 
Note that, with LNSR, noise injection to the lower layers usually leads to relatively higher accuracy and stability, implying that LNSR may be more effective when it affects both the lower and higher layers of the network. 


\subsection{Relationship to Previous Noise-based Approaches}
Our method is related to SMART \cite{Jiang2020SMARTRA}, FreeLB \cite{Zhu2020FreeLBEA} and R3F \cite{Aghajanyan2020BetterFB}. As is mentioned before, most of these approaches employ adversarial training strategies to improve the robustness of BERT fine-tuing. SMART solves supremum by using an adversarial methodology to achieve the largest KL divergence with an $\epsilon$-ball, FreeLB optimizes a direct adversarial loss $\mathcal{L}_{FreeLB}(\theta)=\text{sup}_{\Delta\theta:|\Delta\theta|\leq\epsilon} \mathcal{L}(\theta+\Delta\theta)$ through iterative gradient ascent steps, while R3F removes the adversarial nature of SMART and optimize the smoothness of the whole model directly. 

Compared with this sort of adversarial based algorithms, our method is easier to implement and provides a relatively rigorous theoretical guarantee. The design of layer-wise regularization is sensible that it exploits the characteristics of hierarchical representations in modern deep neural networks. Studies in knowledge distillation have shown similar experience that imitating through middle layer representations \cite{adriana2015fitnets,zagoruyko2016paying} performs better than aligning the final outputs \cite{hinton2015distilling}. Moreover, LNSR allows us to use different regularization weights for different layers (we use fixed weight 1 on all layers in this paper). We will leave the exploitation in future work. 

\section{Conclusion}
In this paper, we propose the Layer-wise Noise Stability Regularization (LNSR) as a lightweight and effective method to improve the generalizability and stability when fine-tuning BERT on few training samples. 
Our proposed LNSR method is a general technique that improves model output stability while maintaining or improving the original performance. Furthermore, we provide a theoretically analysis of the relationship of our model to the Lipschitz continuity and Tikhonov  regularizer. Extensive empirical results show that our proposed method can effectively improve the generalizability and stability of the BERT model. 
\section{Acknowledgements}
Hang Hua would like to thank Jeffries for supporting his research.

\bibliography{anthology,acl2020}
\bibliographystyle{acl_natbib}
\clearpage
\appendix
\section{Experimental Details}
\label{sec:experimental}
\begin{figure*}
\centering 
\begin{minipage}[t]{1.0\linewidth}
\centering
\includegraphics[width=0.99\linewidth]{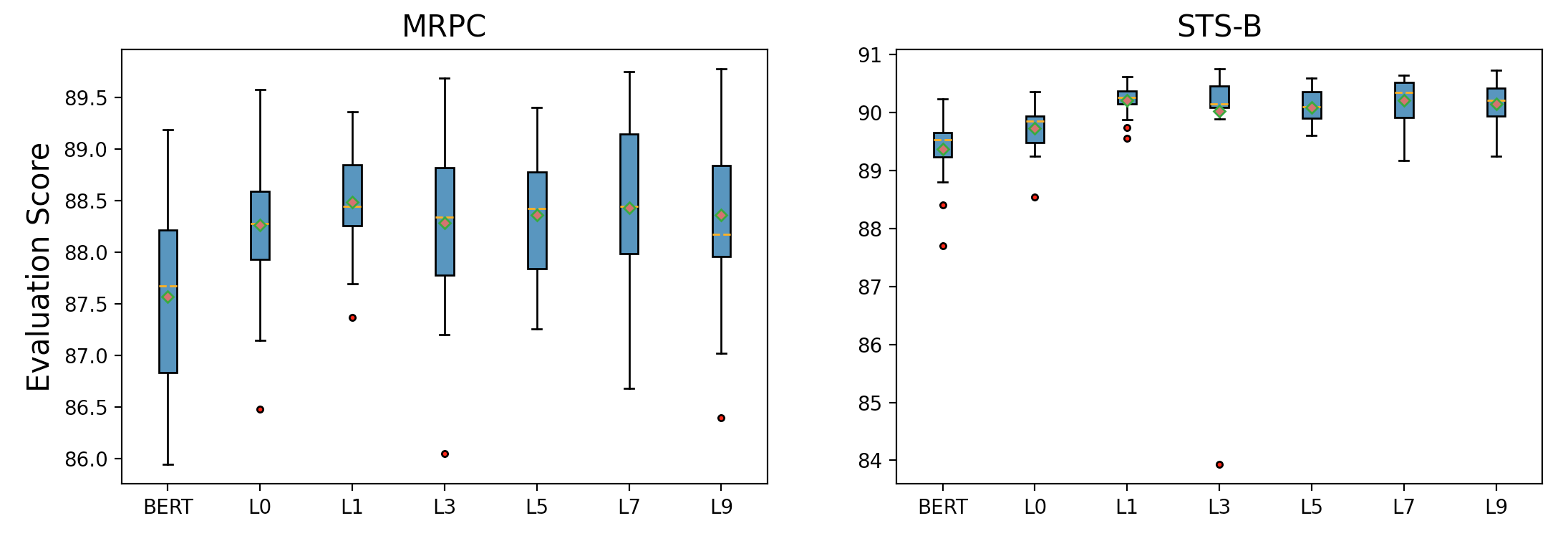}
\end{minipage}

\begin{minipage}[t]{1.0\linewidth}
\centering
\includegraphics[width=0.99\linewidth]{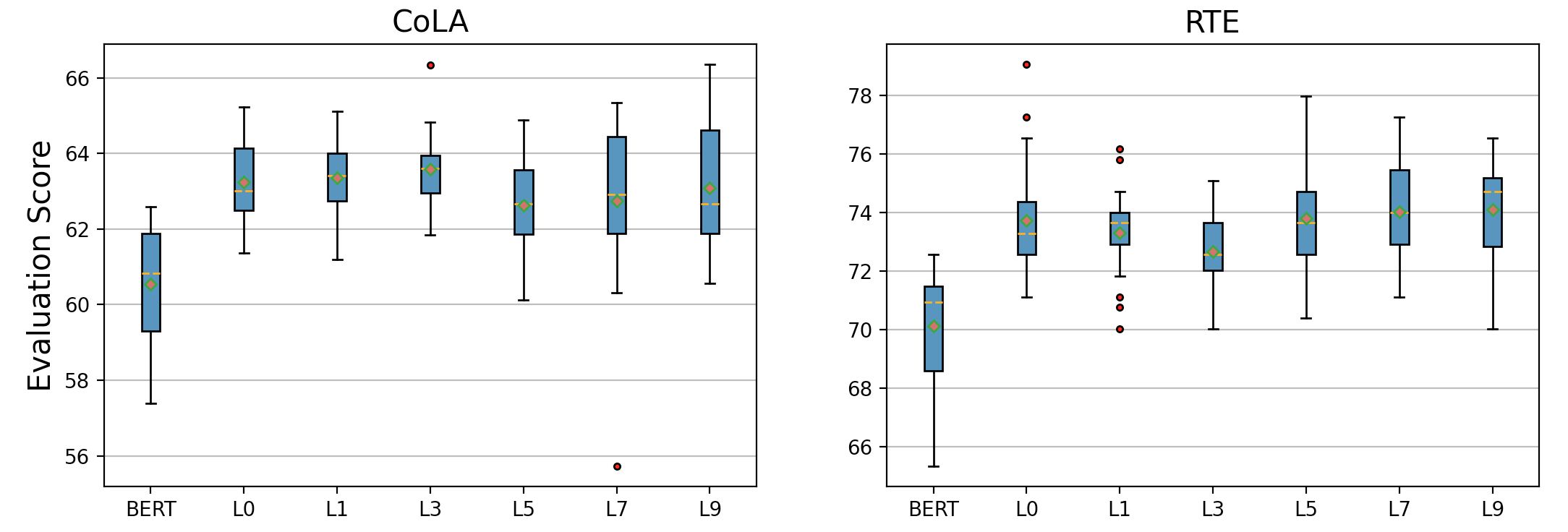}
\end{minipage}
\caption{Performance distribution box plot of each model on the four tasks across 25 random seeds. }
\label{fig:sens}
\end{figure*}

The model we use for experiments in section 4 is the standard BERT large model with 24 layers staked Transformers \cite{Vaswani2017AttentionIA} encoder, 1024 hidden size, and 16 self-attention heads. We initialize the pre-trained part of the model with BERT-Large-Uncased-Whole-Word-Masking weight. The final layer is a classification layer with 2048 parameters which contains $0.0006\%$ of the total number of parameters in the model. We initialize the last layer with $\mathcal{N}(0,0.02^2)$ and each bias is 0. For the position of noise injection, we uniformly chose the first layer as the noise regularization start point. In the sensitivity to the position of noise injection analysis section, we also try injecting noise from the different layers as is shown in Figure~\ref{fig:sens}. As for the baseline model Mixout, we use the code from the Github repository \url{https://github.com/bloodwass/mixout.git}. The other baseline models are implemented by ourselves.

Table~\ref{tab:datasets} summarizes dataset statistics used in this work. We use the standard GLUE benchmark datasets downloaded from \url{https://gluebenchmark.com/tasks}.

\newpage
\section{Other Experimental Reports}
We also report the maximum value we get during fine-tuning BERT with our proposed LNSR regularizer among a large number of random seeds and several noise injection position, since the maximum value can also reflect the ability of the learning algorithm to reach an optimal point. The results are shown in Table~\ref{tab:maxv}, and we can see that on some tasks, fine-tuning BERT with LNSR is even competitive with fine-tuning state-of-the-art models which adopt more powerful modern architectures and pre-training strategies. 

\begin{center}
\begin{table*}[t!]
  \begin{tabular}{lllllll}
    \toprule
    \multirow{2}{*}{} && RTE & MRPC & CoLA & STS-B \\
\midrule
    &Task& NLI & Paraphrase & Acceptability & Similarity \\
    & Metrics & Accuracy & $\frac{\text{Accuracy+F1}}{2}$ & Matthews corr & Pearson/Spearman corr\\
    & $\#$ of labels & 2 & 2 & 2 & 1\\
    & $\#$ of training samples & 2.5k & 3.7k & 8.6k & 7k\\
    & $\#$ of validation samples & 276 & 408 & 1k & 1.5k\\
    & $\#$ of test samples & 3k & 1.7k & 1k & 1.4k\\
    \bottomrule
  \end{tabular}
    \caption{The summarization of the datasets used in this work.}
    \label{tab:datasets}
\end{table*}
\end{center}

\begin{table*}[t!]
\begin{center}
  \begin{tabular}{lllllll}
    \toprule
    \multirow{2}{*}{} && RTE & MRPC & CoLA & STS-B \\
\midrule
    &BERT \cite{Devlin2019BERTPO}& 70.4 & 88.0 & 60.6 & 90.0 \\
    &BERT \cite{Phang2018SentenceEO}& 70.0 & \bf{90.7} & 62.1 & 90.9 \\
    &LNSR (ours)& $\bf{79.1}$ & $90.4$ & $\bf{68.1}$ & $\bf{91.0}$ \\
    \midrule
    &XLNet \cite{Yang2019XLNetGA} & $83.8$ & $89.2$ & $63.6$&$91.8$ \\
    &RoBERTa \cite{Liu2019RoBERTaAR} & $86.6$ & $90.9$ &$68.0$& $92.4$  \\
    &ALBERT \cite{Lan2020ALBERTAL} & $89.2$ & $90.9$ &$71.4$ & $93.0$  \\
    \bottomrule
  \end{tabular}
  \end{center}
    \caption{We report the maximum value we get when fine-tuning the LNSR model on different noise injection position and random seeds on the four tasks. On some tasks, BERT (standard BERT-large-uncased \cite{Devlin2019BERTPO}) with LNSR even become competitive with some newly proposed powerful models (bottom rows)}.
    \label{tab:maxv}
\end{table*}

\end{document}